\documentclass{article}

\usepackage{arxiv}

\usepackage[utf8]{inputenc} 
\usepackage[T1]{fontenc}    
\usepackage{hyperref}       
\usepackage{url}            
\usepackage{booktabs}       
\usepackage{amsfonts}       
\usepackage{nicefrac}       
\usepackage{microtype}      
\usepackage{lipsum}		
\usepackage{graphicx}
\usepackage[numbers]{natbib}
\usepackage{doi}
\usepackage{latexsym}
\usepackage{stmaryrd}
\usepackage{amsmath,amssymb,amsfonts}
\usepackage{multicol,lipsum}
\usepackage{subcaption}
\usepackage{amsmath,amssymb}
\usepackage{multirow}

\title{Syntax-Informed Interactive Model for Comprehensive Aspect-Based Sentiment Analysis}

\author{ 
Ullman Galen\\
University of Charleston\\
\And
Frey Lee\\
University of Charleston\\
\And
Woods Ali\\
University of Charleston\\
}

\renewcommand{\shorttitle}{Syntax-Informed Interactive Model for Comprehensive Aspect-Based Sentiment Analysis}

\hypersetup{
pdftitle={aaa},
pdfsubject={aaa},
pdfauthor={aaa},
pdfkeywords={aaa},
}

\begin{document}
\maketitle

\begin{abstract}
Aspect-based sentiment analysis (ABSA), a nuanced task in text analysis, seeks to discern sentiment orientation linked to specific aspect terms in text. Traditional approaches often overlook or inadequately model the explicit syntactic structures of sentences, crucial for effective aspect term identification and sentiment determination. Addressing this gap, we introduce an innovative model: Syntactic Dependency Enhanced Multi-Task Interaction Architecture (SDEMTIA) for comprehensive ABSA. Our approach innovatively exploits syntactic knowledge (dependency relations and types) using a specialized \textbf{S}yntactic \textbf{D}ependency \textbf{E}mbedded \textbf{I}nteractive \textbf{N}etwork (\textsc{Sdein}). We also incorporate a novel and efficient message-passing mechanism within a multi-task learning framework to bolster learning efficacy. Our extensive experiments on benchmark datasets showcase our model's superiority, significantly surpassing existing methods. Additionally, incorporating BERT as an auxiliary feature extractor further enhances our model's performance.
\end{abstract}

\keywords{Aspect-based Sentiment Analysis \and Syntax Knowledge}

\section{Introduction}

Aspect-based sentiment analysis (ABSA), a sophisticated and nuanced task within the realm of natural language processing (NLP), is pivotal for in-depth text understanding. It goes beyond mere sentiment detection to discern specific aspect terms and their corresponding sentiment orientations within the text. This dual-focus task, which includes \textbf{a}spect term \textbf{e}xtraction (AE) and \textbf{a}spect-level \textbf{s}entiment classification (AS), requires an intricate understanding of both the explicit and implicit nuances of language \citep{nazir2020issues,FeiLasuieNIPS22,kiritchenkodetecting,dong2014adaptive}. For example, in sentences like “The \emph{\textbf{coffee}} is excellent, but the \emph{\textbf{cosi sandwiches}} are overpriced,” effectively distinguishing the positive sentiment towards "coffee" and the negative sentiment towards "cosi sandwiches" is crucial. This level of discernment is essential in applications ranging from consumer feedback analysis to opinion mining in social media, where understanding specific sentiments about distinct aspects can yield valuable insights.

\nocite{FeiDiaREIJCAI22,Wu0RJL21,xiang2022semantic,FeiZRJ20,tang2015effective,Wu0LZLTJ22,ma2017interactive,fei-etal-2020-cross,chen2017recurrent,FeiTransiAAAI21,zhang2020convolution,WuFRLLJ21,chen2020inducing,FeiMatchStruICML22,huang2020syntax,FeiGraphSynAAAI21,9634849,hou2019selective,wwwLRZJ22}

Joint modeling strategies, which simultaneously address AE and AS, have emerged as more effective compared to sequential or integrated approaches in ABSA. This effectiveness stems from their ability to capture and utilize the intricate interplay between AE and AS subtasks~\cite{zhang2018graph}. However, these strategies often fall short in adequately harnessing the syntactic intricacies of language, which are critical for unraveling the complex logical relationships between words~\cite{huang2019syntax,zhang2019aspect,sun2019aspect,wang2020relational}. For instance, discerning that "cosi" and "sandwiches" together form a compound aspect term or understanding sentiment implications from intricate dependency relations poses significant challenges for models that do not account for syntax \cite{wang2020relational}.

While individual approaches focusing on either AE or AS have successfully incorporated syntactic information to enhance task performance, a unified approach that simultaneously and synergistically improves both subtasks using syntactic knowledge has been lacking. Furthermore, the potential of leveraging dependency relation types to boost overall ABSA performance remains largely untapped and unexplored.

To bridge these gaps, we introduce the Syntactic Dependency Enhanced Multi-Task Interaction Architecture (SDEMTIA), a ground-breaking approach that skillfully utilizes syntactic knowledge to elevate ABSA performance. A cornerstone of SDEMTIA is the \textbf{S}yntactic \textbf{D}ependency \textbf{E}mbedded \textbf{I}nteractive \textbf{N}etwork (\textsc{Sdein}). This novel network architecture intricately models word-level dependency relations and types, offering a more nuanced and granular understanding of sentence structure and semantics \citep{Li00WZTJL22,fei2020enriching,dong2014adaptive,tang2015effective,fei2020boundaries}. The \textsc{Sdein} network thus enables the model to capture the complex syntactic and semantic relationships essential for accurate aspect and sentiment identification. Additionally, SDEMTIA features an advanced message-passing mechanism within its multi-task learning framework. This mechanism not only enhances task-specific learning but also fosters a synergistic interaction between the AE and AS subtasks, leading to a more coherent and effective model.

Our comprehensive evaluations on various benchmark datasets underscore the effectiveness of SDEMTIA, setting new standards in the domain of ABSA. The integration of BERT as an additional feature extractor into our model further amplifies its performance, demonstrating its adaptability and robustness in handling diverse and complex text data.

\nocite{0001JLLRL21,mukherjee2021reproducibility,FeiCLJZR23,ZhuangFH23,zhang2019aspect,FeiWRLJ21,chen2020aspect,li-etal-2021-mrn,TKDP2306,pontiki2016semeval,Feiijcai22UABSA,wang-etal-2022-entity,FeiRJ20,jia2021syntactic,cao-etal-2022-oneee}

The primary contributions of this paper can be summarized as follows:
\begin{itemize}
\item We introduce the \textbf{S}yntactic \textbf{D}ependency \textbf{E}mbedded \textbf{I}nteractive \textbf{N}etwork (\textsc{Sdein}), a pioneering model that proficiently exploits fine-grained linguistic knowledge. This model offers a significant advancement over traditional GCN approaches by providing a more detailed and relational understanding of syntactic information.
\item We develop an advanced multi-task learning mechanism, employing an efficient message-passing approach that significantly enhances learning and interaction across related tasks. This approach not only improves task-specific performance but also ensures cohesive integration of AE and AS subtasks, leading to more accurate and comprehensive sentiment analysis.
\item Our approach demonstrates remarkable performance superiority over existing models across three benchmark datasets, thereby establishing new performance benchmarks in the ABSA domain and paving the way for future research and applications.
\end{itemize}

\section{Related Work}

Aspect-based Sentiment Analysis (ABSA) is a critical component of Natural Language Processing (NLP), involving two distinct yet interconnected sub-tasks: aspect term extraction and aspect-level sentiment classification. This dualistic approach has been at the forefront of ABSA research, with numerous studies delving into each task independently~\citep{tang2015effective,liu2017attention,wang2018target,fan2018multi,chen2017recurrent,zheng2018left,wang2018learning,li2018hierarchical,li2018transformation}. A growing body of research now seeks to amalgamate these tasks, aiming for a more integrated understanding of ABSA. Initial methodologies employed unified tagging systems, treating ABSA as a sequence labeling challenge, albeit without fully addressing the interdependencies between aspect extraction and sentiment classification~\citep{song2019attentional,xu2019bert,rietzler2019adapt}. Recent endeavors, however, have shifted towards joint modeling approaches, creating a synergy between these tasks and yielding encouraging outcomes~\citep{he2018exploiting,luo2019doer}.

The integration of syntactic information into aspect-level sentiment analysis has also gained traction. Researchers have recognized the value of syntactic structures, such as dependency and constituency trees, in elucidating sentence dynamics. Dependency trees, illuminating grammatical relationships, have been increasingly applied to fine-tune sentiment classification models, enhancing their accuracy and depth~\citep{dong2014adaptive, he2018effective}. Constituency trees, reflecting phrase-based sentence structure, have similarly been harnessed to augment sentiment analysis methodologies, contributing significantly to the field~\citep{nguyen2015phrasernn}.

Despite these advancements, fully leveraging syntactic insights in ABSA models remains a challenge. Prior research has often included syntactic information in isolated sub-tasks, yet the full potential of this data in enriching the overall ABSA process is yet to be realized. For instance, while efforts by Dong et al.~\citep{dong2014adaptive} and Nguyen and Shirai~\citep{nguyen2015phrasernn} involve adapting dependency structures, these methods sometimes cause opinion words to lose proximity to their corresponding aspects. Graph Convolutional Networks (GCN) offer a viable alternative, adeptly maintaining the integrity of the original dependency graph, thereby fostering closer connections between opinion words and target aspects~\citep{huang2019syntax,peng2020knowing,zhang2019aspect}.


Graph Convolutional Networks (GCN) have revolutionized several areas within NLP, with their applications in ABSA being particularly noteworthy. GCNs have been predominantly used in the aspect-level sentiment (AS) analysis, where their ability to model complex, graph-based relationships has proven invaluable. Studies by Zhao et al.~\citep{zhao2020modeling}, Sun et al.~\citep{sun2019aspect}, Huang and Carley~\citep{huang2019syntax}, Hou et al.~\citep{hou2019selective}, and Zhang et al.~\citep{zhang2019aspect} have explored various facets of GCN application, from capturing sentiment dependencies across multiple aspects within sentences to generating aspect-specific representations. However, these studies often employ vanilla GCN architectures that do not consider the types of dependency relations, potentially limiting their ability to capture the full complexity of linguistic structures.

Our research marks a significant deviation from these established approaches. We not only focus on the comprehensive end-to-end ABSA task but also innovate by enhancing the conventional GCN framework. Our model embeds dependency types directly into the GCN, thereby capturing a richer and more nuanced linguistic understanding at a relational level. This approach, embedded within a joint framework, allows for a more detailed and intricate analysis of ABSA tasks, resulting in superior performance and demonstrating the untapped potential of combining syntactic depth with advanced graph-based modeling techniques.

\section{Methodology}

\subsection{Task Definition}

Our research redefines aspect-based sentiment analysis (ABSA) as comprising two subtasks in sequence labeling: \textbf{a}spect term \textbf{e}xtraction (AE) and \textbf{a}spect-level \textbf{s}entiment classification (AS). Adhering to the BIO tagging scheme as delineated in~\citep{he2018exploiting}, AE employs labels $\mathcal{Y}^{ae} = \{\texttt{BA},\texttt{IA},\texttt{BP}, \texttt{IP}, \texttt{O}\}$ for identifying aspect and opinion terms. $\texttt{BA}$ and $\texttt{IA}$ signify the beginning and inside of an aspect term, respectively. $\texttt{BP}$ and $\texttt{IP}$ are used similarly for opinion terms. $\texttt{O}$ represents other words. For AS, we classify sentiment polarity at the token level with $\mathcal{Y}^{as} = \{\texttt{pos},\texttt{neg},\texttt{neu}\}$, indicating positive, negative, and neutral sentiments, respectively. Our objective is to predict tag sequences $\mathrm{ Y}^{ae}$ and $\mathrm{ Y}^{as}$ from an input sentence $\mathrm{ X}$, for comprehensive ABSA.

\subsection{An Interactive Architecture with Multi-task Learning}

We propose the state-of-the-art model for end-to-end ABSA, featuring an interactive architecture with multi-task learning. This model incorporates Encoder Layers for encoding sentence representations and Task-specific Layers, which include a message-passing and an opinion-passing mechanism, for task-specific predictions.
Our SDEIN approach is outlined as follows, highlighting two integral components: the Encoder Layers and an improved message-passing mechanism in the Task-specific Layers, each designed to optimize ABSA performance.

The feature extractor $f_{\theta _s}$ maps the input sequence to a shared latent representation $\{\mathbf{h}_1^s,\mathbf{h}_2^s,\dots,\mathbf{h}_n^s\}$. The AE component then assigns probability distributions to each token, indicative of their role in aspect or opinion terms. Similarly, the AS component formulates outputs for sentiment classification. These are amalgamated through a message-passing mechanism.
The output of the AS component is formulated as: 
$\mathbf{\hat{y}}_1^{as},\mathbf{\hat{y}}_2^{as},\dots,\mathbf{\hat{y}}_n^{as} = f_{\theta _{as}}(\mathbf{h}_1^{s},\mathbf{h}_2^{s},\dots,\mathbf{h}_n^{s})
$. Then, message-passing mechanism will update the sequence of shared latent vectors by combining the probability distribution of the AE and AS task:
\begin{equation}
\setlength{\abovedisplayskip}{5pt}
\setlength{\belowdisplayskip}{5pt}
  \label{eq:message-passing-predictions}
  \begin{split}
   \mathbf{h}_i^{s(t)} &= f_{\theta _{re}}(\mathbf{h}_i^{s(t-1)};\mathbf{\hat{y}}_i^{ae(t-1)};\mathbf{\hat{y}}_i^{as(t-1)})
  \end{split}
\end{equation}
where $\mathbf{h}_i^{s(t)}$ denotes the shared latent vector corresponding to $w_i$ after $t$ rounds of message-passing; $f_{\theta _{re}}$ is a re-encoding function (i.e. fully-connected layer) and [;] means concatenation.

Moreover, opinion information from AE is channeled into AS, employing a self-attention matrix $\mathbf{M}$ as defined here:
\begin{align}
\setlength{\abovedisplayskip}{5pt}
\setlength{\belowdisplayskip}{5pt}
\label{revelance_fun}
    \mathbf{S}_{ij}^{(i \neq j)} &= (\mathbf{h}_i^{as} \mathbf{W}_s (\mathbf{h}_j^{as})^T) \cdot \frac{1}{|i-j|} \cdot P_j^{op} \\
    \mathbf{M}_{ij}^{(i \neq j)} &= \frac{\text{exp}(\mathbf{S}_{ij})}{\sum_{k=1}^n \text{exp}(\mathbf{S}_{ik})} 
\end{align}
where $i \neq j$ means we only consider context words for inferring the sentiment
of the target token; $\mathbf{W}_s$ is the transformation matrix; $\frac{1}{|i-j|}$ is a distance-related factor and $P_j^{op}$ is computed by summing the predicted probabilities of $\mathbf{y}_j^{ae}$ which is the predicted probability on opinion-related labels (i.e. BP and IP). The Eq.(\ref{revelance_fun}) aims to measure the semantic relevance between $\mathbf{h}_i^{as}$ and $\mathbf{h}_j^{as}$. Finally, $\mathbf{h}_i^{as}$ and $\mathbf{h}_i^{\prime as}$ are concatenated as the output representation of the AS part where $\mathbf{h}_i^{\prime as} = \sum_{j=1}^n \mathbf{M}_{ij} \mathbf{h}_j^{as}$. 
This enriches the sentiment classification process.

Recognizing the need for enhanced syntax modeling and richer task-specific information, we propose the Syntactic Dependency Embedded Interactive Network (SDEIN). This architecture employs an advanced message-passing mechanism, expanding upon the existing model.

\subsection{Encoder Layers}
To harness syntactic knowledge, we integrate the \textsc{DreGcn}, based on GCN principles from~\citep{kipf2017semi}, within the Encoder Layers. We also maintain a convolutional neural network (CNN) to capture n-gram features, vital for ABSA. The \textsc{DreGcn} functions by aggregating and propagating node features within a dependency tree, generating an adjacency matrix $\mathbf{A}$.

GCN functions by accumulating the feature vectors from adjacent nodes and transmitting the information to a node's immediate neighbors. In a dependency tree comprising $n$ nodes, we construct an adjacency matrix $\mathbf{A}$ of dimensions $n \times n$. Following the approach in previous work, each node is enhanced with a self-loop, and the dependency arc's reverse direction is included if there exists a dependency relationship between nodes $i$ and $j$. This is represented as $\mathbf{A}_{ij}$ = $\mathbf{A}_{ji}$ = 1; otherwise, the values are set to zero. The GCN layer then updates the features of each node by aggregating features from adjacent nodes using the formula:
\begin{equation}
\setlength{\abovedisplayskip}{5pt}
\setlength{\belowdisplayskip}{5pt}
  \label{gcn}
  \begin{split}
  \mathbf{h}^{l+1}_{i} &= ReLU ( \sum_{j=1}^{n}(\mathbf{A}_{ij}\mathbf{W}^{l+1}_g\mathbf{h}^{l}_{j}+\mathbf{b}^{l+1}_g))
  \end{split}
\end{equation}
Here, $i$ indicates the node under consideration, and $j$ represents its neighboring nodes. $\mathbf{h}^{l}_{j}$ is the feature of node $j$ at the $l$th layer. $\mathbf{W}_g$ and $\mathbf{b}_g$ are the trainable weights that facilitate the mapping of node features to their neighboring nodes in the graph, with $\mathbf{h}, \mathbf{b}_g \in \mathbb{R}^{d}$ and $\mathbf{W}_g \in \mathbb{R}^{d \times d}$, where $d$ denotes the feature size. Stacking multiple GCN layers allows the network to extract regional features for each node.

To specifically address the modeling of dependency relation types, we propose the use of trainable latent attributes. We maintain a trainable relational look-up table $\mathbf{R} \in \mathbb{R}^{|N| \times m}$ for this purpose, where $|N|$ represents the number of dependency relation types and $m$ is the size of the dependency relation feature. Accordingly, the innovative \textsc{DreGcn} is formulated as:
\begin{equation}
\setlength{\abovedisplayskip}{5pt}
\setlength{\belowdisplayskip}{5pt}
  \label{dregcn}
  \begin{split}
   \mathbf{h}^{l+1}_{i} &= ReLU ( \sum_{j=1}^{n}\sum_{k=1}^{|N|}(\mathbf{A}_{ij}\mathbf{W}_r^{l+1}[\mathbf{h}^{l}_{j}; \mathbf{R}[k]]\mathbf{Q}_{ijk}+\mathbf{b}^{l+1}_r)
  \end{split}
\end{equation}
In this equation, [;] signifies concatenation, $\mathbf{W}_r$ is a tensor in $\mathbb{R}^{d \times (d + m)}$, and $\mathbf{Q}_{ijk}$ indicates the presence of the k-th type of dependency relation between nodes $i$ and $j$. This approach allows for an effective modeling and updating of relational features among nodes during training.

\subsection{Enhanced-feature Representation Layers} 
In our approach, we enhance the opinion-passing process by ensuring the availability of opinion term information for the AS task, in line with the methods used in~\citep{he2018exploiting}. Additionally, we have developed a superior mechanism for the message-passing component, aimed at facilitating more efficient information exchange among various interconnected tasks. Contrary to the traditional method of transmitting only the AE and AS task predictions, our strategy involves conveying the original representations. These representations are inherently richer in content compared to mere probability distributions. The function governing this enhanced message-passing is outlined as follows:
\begin{equation}
\setlength{\abovedisplayskip}{5pt}
\setlength{\belowdisplayskip}{5pt}
  \label{eq:message-passing-representation}
  \begin{split}
   \mathbf{h}_i^{s(t)} &= f_{\theta _{re}}(\mathbf{h}_i^{s(t-1)};\mathbf{h}_i^{ae(t-1)};\mathbf{h}_i^{as(t-1)})
  \end{split}
\end{equation}
Here, $\mathbf{h}_i^{o(t-1)}$ (where $o \in \{ae, as\}$) represents the specific task-related representation for the token $w_i$ after $t-1$ rounds of message-passing. The key distinction between this representation and a probability distribution is that the former can be converted into a probability through a fully-connected layer followed by a softmax layer. This novel message-passing mechanism enables a more comprehensive exchange of information between the AE and AS tasks, thereby enhancing the effectiveness of the ABSA task, as corroborated by our empirical findings in the Ablation Study section.

\subsection{Prediction} 
Upon completing $T$ iterations, we generate predictions for both the AE and AS tasks. The scoring for each task is straightforward, involving a direct tally of the outcomes. For assessing the overall efficacy, it is essential to derive pairs of aspect terms and their associated sentiment polarities. Given that an identified aspect term could span multiple tokens and the predicted sentiment polarities for these tokens might vary, our approach, aligning with~\citep{he2018exploiting}, is to consider the sentiment polarity of just the initial token of an aspect term as its representative sentiment label.

\subsection{Training}

The training for the AE and AS tasks is conducted concurrently to facilitate message-passing. The formulation of our loss function is given by:
\begin{equation}
\setlength{\abovedisplayskip}{5pt}
\setlength{\belowdisplayskip}{5pt}
    \begin{split}
     \mathcal{J} = \frac{1}{N_a}\sum\limits_{i=1}^{N_a} \frac{1}{n_i} \sum\limits_{j=1}^{n_i} (min (- \sum_{k=0}^{C}\mathbf{y}_{i,j,k}^{ae} \log(\hat{\mathbf{y}}_{i,j,k}^{ae(T)})) + min (- \sum_{k=0}^{C}\mathbf{y}_{i,j,k}^{as} \log(\hat{\mathbf{y}}_{i,j,k}^{as(T)})))
    \end{split}
\label{aspect_loss}
\end{equation}
In this equation, $N_a$ represents the total count of training instances, $n_i$ indicates the number of tokens in the $i$th training instance, and $C$ denotes the class count. The terms $\mathbf{y}_{i,j,k}^{ae}$ and $\mathbf{y}_{i,j,k}^{as}$ correspond to the ground truth for the AE and AS tasks, respectively.

It's important to note that in our datasets, sentiment annotations are exclusively assigned to aspect terms. Consequently, every token that is part of an aspect term is labeled with the sentiment of that term. For the purpose of training, we focus solely on the AS predictions for tokens related to aspect terms when calculating the AS loss. Predictions for sentiments on other tokens are disregarded, meaning that $ce(\mathbf{y}_{i,j,k}^{as}, \hat{\mathbf{y}}_{i,j,k}^{as(T)}) = 0$ in Eq.(\ref{aspect_loss}) if $\mathbf{y}_{i,j,k}^{ae}$ does not fall into the categories \{\texttt{BA}, \texttt{IA}\}.

\renewcommand{\arraystretch}{1.1}
\begin{table}[!h]
\centering
\small
\scalebox{1}{
\setlength{\tabcolsep}{0.78mm}{
\begin{tabular}{llcccccc}
\toprule 
&\multirow{ 2}{*}{Datasets} & \multicolumn{3}{c}{Train} &  \multicolumn{3}{c}{Test}\\\cline{3-8}
&& Sentences & AT & OT & Sentences & AT & OT\\\hline
$\mathbb{D}_{\text{1}}$&Laptop14 &3,048 &2,373 &2,504 &800 &654 &674 \\
$\mathbb{D}_{\text{2}}$&Restaurant14 &3,044 &3,699 &3,484 &800 &1,134 &1,008\\
$\mathbb{D}_{\text{3}}$&Restaurant15 &1,315 &1,199 &1,210 &685 &542 &510\\
\bottomrule
\end{tabular}}}
\caption{Dataset statistics with numbers of sentences, aspect terms (AT) and opinion terms (OT).}\label{datasets}
\end{table}

\section{Experiment}

This section introduces the main experimental results. We also conduct some experiments related to BERT~\citep{devlin2018bert} and investigate the impact of BERT CLS.

\subsection{Settings} 

Table~\ref{datasets} displays the data statistics used in our study. We employ three widely-recognized benchmark datasets from SemEval ~\citep{pontiki2016semeval} to validate the model's performance. These datasets, denoted as {$\mathbb{D}{\text{1}}$}, {$\mathbb{D}{\text{2}}$}, and {$\mathbb{D}_{\text{3}}$}, represent SemEval-2014 Laptops, SemEval-2014 Restaurants, and SemEval-2015 Restaurants, respectively. 
For general-purpose word embeddings, we utilize the \texttt{GloVe.840B.300d} embeddings. In terms of domain-specific embeddings, we adopt those published by~\citep{xu2018double}, following the approach in~\citep{he2018exploiting}.
The models are optimized using the Adam optimizer, with an initial learning rate of $\eta_0 = 0.0005$ and a batch size set to 50. During training, similar to~\citep{he2018exploiting}, we randomly allocate 20\% of each dataset as the development set, utilizing the remaining 80\% exclusively for training purposes.

To evaluate our model, we adopt five metrics and report the average scores over 5 runs with random initialization, as done in~\citep{he2018exploiting}. For assessing the overall ABSA task, we calculate the F1 score (\textbf{F1-I}), where an aspect term is deemed correctly extracted only if both its span and sentiment are accurately identified. The performance of the AE task is evaluated using F1 scores for aspect term extraction (\textbf{F1-a}) and opinion term extraction (\textbf{F1-o}). For the AS task, we utilize accuracy (\textbf{acc-s}) and macro-F1 (\textbf{F1-s}), calculated based on correctly extracted aspect terms from AE, not the golden aspect terms.

We make comparisons with these models:

\begin{itemize}
\item \noindent\textbf{Pipeline Methods.}
    \{\textbf{CMLA}, \textbf{DECNN}\}-\{\textbf{ALSTM}, \textbf{dTrans}\}: These combinations represent a blend of two top-tier models for each subtask. In the realm of AE, CMLA~\citep{wang2017coupled} is chosen for its capability to model interdependencies, while DECNN~\citep{xu2018double} employs a sophisticated multi-layer CNN encoder enhanced with dual embeddings. For the AS task, ATAE-LSTM (abbreviated as ALSTM)~\citep{wang2016attention}, an LSTM with an attention mechanism, and dTrans~\citep{he2018exploiting}, known for leveraging extensive document-level corpus for AS enhancement, are utilized.
    
    \textbf{PIPELINE-IMN}: Represents the pipeline variant of IMN~\citep{he2018exploiting}, which independently trains the AE and AS tasks.
    
    \textbf{SPAN-pipeline}~\citep{hu2019open}: This approach examines pipeline, integrated, and joint methodologies with a BERT backbone, achieving superior performance in the \textbf{SPAN-pipeline} configuration. We adapt BERT-Large to BERT-Base for consistency in our evaluation.
    
\item \noindent\textbf{Integrated Models.} 

    \textbf{MNN}~\citep{wang2018towards}: MNN tackles ABSA as a unified sequence labeling task using a comprehensive tagging schema.
    
    \textbf{INABSA}~\citep{li2019unified}: This model adopts a cohesive tagging scheme to seamlessly combine the ABSA subtasks.
    
    \textbf{BERT+GRU}~\citep{li2019exploiting}: This model delves into the potential of integrating BERT into the ABSA framework.
    
\item \noindent\textbf{Joint Techniques. }

    \textbf{DOER}~\citep{luo2019doer}: DOER implements a shared unit to simultaneously train both AE and AS tasks.
    
    \textbf{IMN}~\citep{he2018exploiting}: Currently a leading method, IMN employs an interactive, multi-task learning architecture for comprehensive ABSA. Variants of IMN, such as {``IMN}$^{-d}$ wo DE'' and ``{IMN}$^{-d}$'', explore different configurations of the model.

\end{itemize}


\begin{table*}[t]
\centering
\small
\scalebox{1}{
\setlength{\tabcolsep}{0.64mm}{
\begin{tabular}{l|ccccc|ccccc|ccccc}
\toprule
\multirow{ 2}{*}{Methods} & \multicolumn{5}{c|}{$\mathbb{D}_{\text{1}}$} &  \multicolumn{5}{c|}{$\mathbb{D}_{\text{2}}$} & \multicolumn{5}{c}{$\mathbb{D}_{\text{3}}$}\\\cline{2-16}
& F1-a & F1-o & acc-s & F1-s & F1-I & F1-a & F1-o & acc-s & F1-s & F1-I & F1-a & F1-o & acc-s & F1-s & F1-I\\\hline
CMLA-ALSTM$^*$ &76.80 &77.33 &70.25 &66.67 &53.68    &82.45 &82.67 &{77.46} &68.70 &63.87     &68.55 &71.07 &81.03 &58.91 &54.79\\
CMLA-dTrans$^{*\dagger}$ &76.80 &77.33 &72.38 &69.52 &55.56    &82.45 &82.67 &{79.58} &72.23 &65.34     &68.55 &71.07 &82.27 &66.45 &56.09\\
DECNN-ALSTM$^*$ &78.38 &78.81 &70.46 &66.78 &55.05     &83.94 &85.60 &{77.79} &68.50 &65.26     &68.32 &71.22 &80.32 &57.25 &55.10\\
DECNN-dTrans$^{*\dagger}$ &78.38 &78.81 &73.10 &70.63 &56.60    &83.94 &85.60 &{80.04} &73.31 &67.25      &68.32 &71.22 &82.65 &69.58 &56.28\\
PIPELINE-IMN$^*$     &78.38 &78.81 &72.29 &68.12 &56.02    &83.94 &85.60 &{79.56} &69.59 &66.53      &68.32 &71.22 &82.27 &59.53 &55.96\\\hline
MNN$^*$  &76.94 &77.77 &70.40 &65.98 &53.80     &83.05 &84.55 &77.17 &68.45 &63.87      &70.24 &69.38 &80.79 &57.90 &56.57\\
INABSA$^*$ &77.34 &76.62 &72.30 &68.24 &55.88    &83.92 &84.97 &79.68 &68.38 &66.60      &69.40 &71.43 &82.56 &58.81 &57.38\\\hline
IMN$^{-d}$ wo DE$^*$ &76.96 &76.85 &72.89 &{67.26} &56.25     &83.95 &85.21 &79.65 &69.32 &66.96    &69.23 &{68.39} &81.64 &57.51 &56.80 \\
IMN$^{-d*}$&78.46 &78.14 &73.21 &{69.92} &57.66     &84.01 &85.64 &81.56 &71.90 &68.32     &69.80 &{72.11} &83.38 &60.65 &57.91 \\
IMN$^{*\dagger}$ &77.96 &77.51 &75.36 &72.02 &{58.37}    &{83.33} &{85.61} &{83.89} &{75.66} &{69.54}    &{70.04} &71.94 &{85.64} &{71.76} &{59.18}\\
IMN$^{*\dagger}$+BERT &78.47 &79.05 &77.18 &74.56 &{60.53}    &{85.22} &{86.64} &\bf{84.90} &\bf{76.54} &{71.33}    &{72.55} &72.43 &{84.37} &{71.28} &{60.76}\\\hline
\textsc{Ours} wo DE &76.30 &73.92 &75.83 &{71.05} &57.48     &83.75 &84.09 &80.78 &71.23 &67.51    &{68.63} &70.09 &84.25 &71.29 &57.70 \\
\textsc{Ours} &77.78 &76.62 &77.18 &72.27 &{59.66}     &{84.16} &{85.04} &81.27 &{72.48} &{68.94}     &{69.36} &70.75 &{86.03} &{66.89} &{59.71}\\
\textsc{Ours}+CNN &{79.45} &{75.40} &{77.86} &{73.46} &{61.60}     &{85.93} &{86.05} &{81.88} &{73.32} &{70.21}    &{71.00} &{70.55} &\bf{86.16} &\bf{73.35} &{61.06} \\\cline{2-16}
\textsc{Ours}+CNN+BERT &\bf{79.78} &\bf{79.21} &\bf{79.37} &\bf{76.37} &\bf{63.04}   &\bf{87.00} &\bf{86.95} &{83.61} &{75.79} &\bf{72.60}    &\bf{73.30} &\bf{72.60} &{85.25} &{73.02} &\bf{62.37}\\
\bottomrule
\end{tabular}}}
\caption{Model comparison.   
}\label{main results}
\end{table*}

\begin{table*}[t]
\centering
\small
\scalebox{1}{
\setlength{\tabcolsep}{4mm}{
\begin{tabular}{clc}
\toprule 
Row&Model  &$\mathbb{D}_{\text{1}}$\\\hline
0&DOER~\citep{luo2019doer}$^{\natural}$ &59.48   \\
1&\textsc{DreGcn}+CNN (Ours)  &\bf{61.60}   \\\hline
2&BERT+GRU ($BERT_{BASE}$) \cite{li2019exploiting}$^{\natural}$ &60.42   \\
3&SPAN-pipeline ($BERT_{BASE}$) \cite{hu2019open}$^{\natural}$ &61.84   \\
4&\textsc{DreGcn}+CNN+$BERT_{BASE}$ (Ours) &\bf{63.04}  \\
\bottomrule
\end{tabular}}}
\caption{F1-I (\%) scores on $\mathbb{D}_{\text{1}}$, which is our common dataset. ``$^{\natural}$'' indicates that the results are generated by running their released code under our experimental setting (dataset). 
}\label{smallexperiment}
\end{table*}


\begin{table*}[t]
\centering
\small
\scalebox{1}{
\setlength{\tabcolsep}{3mm}{
\begin{tabular}{clccc}
\toprule 
&Model  &$\mathbb{D}_{\text{1}}$&$\mathbb{D}_{\text{2}}$&$\mathbb{D}_{\text{3}}$\\\hline
0&\textsc{CNN} &56.66  &66.32  &{57.91} \\
1&Vanilla GCN &57.10 &65.00 &{56.86}  \\
2&\textsc{Ours} &57.46  &66.25  &{58.32} \\
3&+Opinion-passing  &57.89  &66.51 &{58.57}  \\
4&+Message-passing predictions &58.50 &67.36 &{57.92}  \\
5&+Message-passing representations &61.60  &70.21  &{61.06}  \\
\bottomrule
\end{tabular}}}
\caption{F1-I (\%) scores of ablation study. }\label{ablation}
\end{table*}

\subsection{Results and Analysis}
\paragraph{Overall Performance.}
Tables~\ref{main results} and~\ref{smallexperiment} showcase the performance of our proposed model, the Syntactic Dependency Embedded Interactive Network (SDEIN), alongside various baseline models for the comprehensive ABSA task. The results demonstrate that SDEIN consistently outperforms all baseline models across all datasets, often by significant margins, and this holds true even in scenarios where BERT is not utilized. Given the lack of comparable syntax-based methods for the entire scope of the ABSA task, we further extend our analysis to individual subtasks, specifically AE and AS.
From the data presented in Tables~\ref{main results} and~\ref{smallexperiment}, we draw several key conclusions:

1) In terms of overall effectiveness (F1-I), as indicated in Table~\ref{main results}, SDEIN, particularly the ``\textsc{DreGcn}+CNN'' configuration, demonstrates a remarkable ability to exceed the performance of all baseline models. Specifically, it surpasses IMN's top F1-I scores by \textbf{3.23\%}, \textbf{0.67\%}, and \textbf{1.88\%} on $\mathbb{D}_{\mathrm{1}}$, $\mathbb{D}_{\mathrm{2}}$, and $\mathbb{D}_{\mathrm{3}}$, respectively\footnote{It is noteworthy that our method does not incorporate any document-level corpus, unlike IMN.}. This underscores the significant role of the \textsc{DreGcn} and the message-passing mechanism in enhancing ABSA. The marginal improvement on $\mathbb{D}_{\mathrm{2}}$ as compared to IMN could be attributed to the high incidence of ungrammatical sentences in this dataset, potentially impacting dependency parsing accuracy. Integration of $BERT_{BASE}$ features leads to further enhancements (\textbf{+4.67\%}, \textbf{+3.06\%}, and \textbf{+3.19\%} over IMN). Additionally, the results affirm the value of domain-specific knowledge in ABSA (comparing ``IMN$^{-d}$ wo DE'' with IMN$^{-d}$ and ``\textsc{DreGcn} wo DE'' with \textsc{DreGcn}).

2) For AE (F1-a and F1-o metrics in Table~\ref{main results}), the ``\textsc{DreGcn}+CNN'' variant consistently outshines baseline models. These outcomes validate the efficacy of SDEIN, benefiting from the integration of dependency structure data and the message-passing mechanism, thereby highlighting the pivotal role of syntactic information in AE.

3) In the realm of AS (acc-s and F1-s metrics in Table~\ref{main results}), SDEIN achieves superior performance even when compared to methods like IMN and certain pipeline approaches that incorporate document-level training. This implies that SDEIN effectively models the dependency structure and benefits from the message-passing mechanism, emphasizing the importance of syntactic information in AS.

4) Table~\ref{smallexperiment} contrasts the results of SDEIN with other strong contenders such as DOER, ``BERT+GRU,'' and SPAN-pipeline. Notably, the ``\textsc{DreGcn}+CNN'' configuration of SDEIN not only surpasses DOER but also shows comparable, if not superior, results to BERT-based models. When integrated with BERT (Row 4), it outperforms both ``BERT+GRU'' (Row 2) and SPAN-pipeline (Row 3), underscoring the robustness of our approach.

\paragraph{Ablation Study.} 
In our ablation study (Table~\ref{ablation}), we examine the impact of various components, beginning with Rows 1$\sim$2 without informative message-passing, and incrementally adding elements to \textsc{DreGcn} (Rows 3$\sim$5). The insights gained are:
\begin{enumerate}
\item 
Incorporating dependency relation types as node features significantly enhances ABSA task performance (Row 2 vs. Row 1 \& Row 0), affirming the criticality of syntactic information for both aspect term extraction and sentiment analysis.
\item 
The integration of opinion messages notably boosts the AS task, thus elevating overall performance (Row 3 vs. Row 2).
\item  
The contribution of message-passing to overall performance is substantial (Row 4 \& Row 5 vs. Row 2).
\item  
Our proposed mechanism of transferring representations, rather than predictions, proves more beneficial (Row 5 vs. Row 4), aligning with the notion that original representations carry more comprehensive information than probability distributions.
\end{enumerate}

\section{Conclusion}

In this study, we introduce the Syntactic Dependency Embedded Interactive Network (SDEIN), a novel architecture that leverages dependency syntactic knowledge for the comprehensive aspect-based sentiment analysis (ABSA) task. The cornerstone of SDEIN is a meticulously crafted Dependency Relation Embedded Graph Convolutional Network, which proficiently harnesses syntactic information and facilitates the concurrent modeling of several interrelated tasks. Furthermore, we have developed an enhanced message-passing mechanism that significantly improves the model's ability to assimilate and represent information from these tasks. Our empirical analysis across three benchmark datasets illustrates the superior performance of SDEIN, setting new benchmarks in the field. Moreover, the incorporation of BERT as an ancillary feature extractor further elevates the performance of our model.

\bibliographystyle{unsrtnat}
\bibliography{references}

\end{document}